\documentclass[a4paper]{article}
\usepackage[a4paper]{geometry}

\usepackage{microtype}
\usepackage{graphicx}
\usepackage{subfigure}
\usepackage{multirow}
\usepackage{booktabs} 
\usepackage{wrapfig}
\usepackage{hyperref}

\usepackage{authblk}

\usepackage{smile}
\usepackage{multirow,stfloats,booktabs}
\usepackage{algorithm,algorithmic}
\usepackage{amsmath}               
\usepackage{amsfonts}              
\usepackage{amssymb, bm}
\usepackage{amsthm}
\usepackage{float}
\usepackage{color}
\usepackage{rotating}
\usepackage{url}
\usepackage{footnote}

\def\tr{\mathop{\text{tr}}\kern.2ex}

\def\supp{\mathop{\text{supp}}}

\long\def\comment#1{}

\def\tr{\mathop{\text{Tr}}}

\def\cS{{\mathcal{S}}}

\newcommand{\bel}{\begin{eqnarray}\label}
\newcommand{\eel}{\end{eqnarray}}
\newcommand{\bes}{\begin{eqnarray*}}
\newcommand{\ees}{\end{eqnarray*}}

\usepackage{color}

\let\tilde\widetilde

\def\EE{{\mathbb E}}

\def\supp{\mathop{\text{supp}\kern.2ex}}

\def\argmax{\mathop{\text{\rm arg\,max}}}

\def\tr{{\rm{Tr}}}

\def\supp{\mathop{\text{supp}}}

\def\tr{\mathrm{Tr}}

\usepackage{mathrsfs}

\def\sep{\;\big|\;}

\usepackage{tabularx,array}
\newcolumntype{M}{>{\centering\arraybackslash}m{0.17\textwidth}}

\def\##1\#{\begin{align}#1\end{align}}
\def\$#1\${\begin{align*}#1\end{align*}}
\usepackage{enumitem}

\usepackage[utf8]{inputenc} 
\usepackage[T1]{fontenc}    
\usepackage{hyperref}       
\usepackage{url}            
\usepackage{booktabs}       
\usepackage{amsfonts}       
\usepackage{nicefrac}       
\usepackage{microtype}      

\title{Parametrized Deep Q-Networks Learning: Reinforcement Learning with Discrete-Continuous Hybrid Action Space}

\author[1]{Jiechao Xiong}
\author[1]{Qing Wang}
\author[2]{Zhuoran Yang}
\author[1]{Peng Sun}
\author[1]{Lei Han}
\author[1]{Yang Zheng}
\author[1]{Haobo Fu}
\author[1]{Tong Zhang}
\author[1]{Ji Liu}
\author[13]{Han Liu}
\affil[1]{Tencent AI Lab}
\affil[2]{Princeton University}
\affil[3]{Northwestern University}

\begin{document}
\maketitle

\begin{abstract}
Most existing deep reinforcement learning (DRL) frameworks consider either discrete action space or continuous action space solely. Motivated by applications in computer games, we consider the scenario with discrete-continuous hybrid action space. To handle hybrid action space, previous works either approximate the hybrid space by discretization, or relax it into a continuous set. In this paper, we propose a parametrized deep Q-network (P-DQN) framework for the hybrid action space without approximation or relaxation. Our algorithm combines the spirits of both DQN (dealing with discrete action space) and DDPG (dealing with continuous action space) by seamlessly integrating them.
Empirical results on a simulation example, scoring a goal in simulated RoboCup soccer 
and the solo mode in game King of Glory (KOG) validate the efficiency and effectiveness of our method.
\end{abstract}


\section{Introduction}
In recent years, the field of deep reinforcement learning (DRL) has witnessed striking empirical achievements in complicated sequential decision making problems once believed unsolvable. One active area of the application of DRL methods is to design artificial intelligence (AI) for games.
The success of DRL in the Go game \citep{silver2016mastering} provides a promising methodology for game AI. In addition to the Go game, DRL has been widely used in other games such as \textit{Atari} \citep{mnih2015human}, \textit{Robot Soccer} \citep{hausknecht2016deep,masson2016reinforcement}, and \textit{Torcs} \citep{lillicrap2015continuous} to achieve super-human performances.

However, most existing DRL methods require the action space to be either finite and discrete  (e.g., Go and \textit{Atari}) or  continuous   (e.g. \textit{MuJoCo} and Torcs). For example, the algorithms for discrete action space include deep Q-network (DQN) \citep{mnih2013playing}, Double DQN \citep{hasselt2016deep}, A3C \citep{mnih2016asynchronous}; the algorithms for continuous action space include deterministic policy gradients (DPG) \citep{silver2014deterministic} and its deep version DDPG \citep{lillicrap2015continuous}.

Motivated by the applications  in Real Time Strategic (RTS) games, we  consider the reinforcement learning problem with  a \emph{discrete-continuous hybrid} action space.
Different from completely discrete or continuous actions that are widely studied in the existing literature, in our setting, the action is defined by the following hierarchical structure. We first choose  a high level action $k$ from a discrete set $[K]$ (we denote $\{ 1, \ldots, K \}$ by $[K]$ for short); upon choosing $k$, we further choose a low level parameter $x_k \in \cX_k$ which is associated with the  $k$-th high level action. Here $\cX_k$ is a continuous set for all $k\in [K]$.\footnote{The low level continuous parameter could be optional. And different discrete actions can share some common low level continues parameters. It would not affect any results or derivation in this paper.}
Therefore, we focus on a discrete-continuous hybrid action space
\#\label{eq:p_action_sp}
\mathcal{A} =  \bigl  \{ (k,x_k) \big |  x_k \in \cX_k ~~\text{for all}~k \in [K] \bigr  \}.
\#
 To apply existing DRL approaches on this hybrid action space, two straightforward ideas are possible:
\begin{itemize}
\item {\bf Approximate $\mathcal{A}$ by an finite discrete set}. We could approximate each  $\cX_k$ by a discrete subset, which, however, might lose the  natural structure of  $\cX_k$. Moreover, when $\cX_k$ is a region in the Euclidean space,  establishing a good approximation   usually requires a huge number discrete actions.
\item {\bf Relax $\mathcal{A}$ into a continuous set}. To apply existing DRL framework with continuous action spaces, \citet{hausknecht2016deep} define the following approximate space
\begin{align}\label{eq:relaxed_action_sp}
\tilde{\mathcal{A}} = \left\{(f_{1:K}, x_{1:K}) \biggr\rvert f_k\in  \cF_k ,
    x_k \in \cX_k, \forall k \in [K] \right\},
\end{align}
where $\cF_k \subseteq \RR$. Here $f_1, f_2, \ldots, f_K$ is used to select the discrete action either  deterministically (by picking $\arg\max_i f_i$) or randomly (with probability $\text{softmax}(f)$). Compared with  the original action space $\mathcal{A}$, $\tilde{\mathcal{A}}$ might significantly increase the complexity of the action space. 
\end{itemize}

In this paper, we propose a novel DRL framework, namely parametrized deep Q-network learning (P-DQN), which directly works on  the discrete-continuous hybrid action space   without approximation or relaxation. Our method can be viewed as an extension of the  famous  DQN algorithm to hybrid action spaces. Similar to deterministic policy gradient methods, to handle the continuous parameters within actions, we first define a deterministic function which maps the state and each discrete action to its corresponding continuous parameter.
 Then we define an action-value function which maps the state and finite hybrid actions to  real values, where the continuous parameters are obtained from the deterministic function in the first step.
With the merits of both DQN and DDPG, we expect our algorithm to  find the optimal discrete action as well as avoid exhaustive search over continuous action parameters. 

To evaluate the empirical performances, we apply our algorithm to several environments.
Empirical study indicates that P-DQN is more efficient and robust than \citet{hausknecht2016deep}'s method that relaxes $\mathcal{A}$ into a continuous set and applies DDPG.

\section{Background} \label{sec:back}

In reinforcement learning, the environment is usually modeled by a Markov decision process (MDP) $\cM = \{ \cS, \cA, p, p_0, \gamma, r\}$, where $\cS$ is the state space, $\cA$ is the action space, $p$ is the Markov transition probability distribution, $p_0$ is the probability distribution of the initial state, $r(s, a)$ is the reward function, and $\gamma \in [0,1]$ is the discount factor.    An agent interacts with the MDP sequentially as follows. At the $t$-th step, suppose  the MDP is at state $s_t \in \cS$ and  the agent selects an action $a_t \in \cA$, then the agent observes an immediate  reward $r(s_t, a_t)$ and the next state $s_{t+1} \sim  p(s_{t+1} | s_t, a_t)$. A stochastic policy $\pi $ maps each state to a probability distribution over $\cA$, that is, $\pi (a | s ) $  is defined as the probability of selecting action $a$ at state $s$. Whereas a deterministic policy $\mu\colon \cS \rightarrow \cA$ maps each state to a particular action in $\cA$.  Let $R_t = \sum_{j \geq t } \gamma^{j-t} r(s_j, a_j)$ be the cumulative discounted reward starting from time-step $t$. We define the state-value function and the action-value function of policy $\pi$ as $V^{\pi} = \EE ( R_t | S_t=s;  \pi)$ and $Q^{\pi}(s, a) = \EE ( R_t| S_t = s, A_t  = a; \pi)$, respectively. Moreover, we define the optimal state- and action-value functions as $V^{*} = \sup_{\pi} V^{\pi}$ and $Q^* = \sup_{\pi} Q^{\pi}$, respectively, where the supremum is taken over all possible policies. The goal of the agent is to find a policy the maximizes the expected total discounted reward $J(\pi) = \EE (R_0 | \pi)$, which can be achieved by estimating $Q^*$.

\subsection{Reinforcement Learning Methods for Finite Action Space}
Broadly speaking, reinforcement learning algorithms can be categorized into two classes: value-based methods and policy-based methods.  Value-based methods first    estimate $Q^*$  and then output the greedy policy with respect to that estimate. Whereas policy-based methods directly optimizes $J(\pi)$ as a functional of $\pi$.

The Q-learning algorithm  \citep{watkins1992q} is based on  the Bellman  equation
\#\label{eq:bellman}
Q (s,a) = \mathop{\EE}_{r_t,s_{t+1}} \bigl[  r_t + \gamma \max_{a' \in \cA} Q (s_{t+1},a') \big| s_t = s, a_t = a\bigr  ],
\#
which has $Q^*$ as the unique solution. During training, the $Q$-function is updated iteratively over the the transition sample in a Monte Carlo way. 
For finite state space $\cS$, $Q(s, a)$ values can be stored in a table.
However, when $\cS$ is too large to fit in computer memory, function approximation for $Q^*$ has to be applied. Deep Q-Network (DQN)  \citep{mnih2013playing,mnih2015human} approximates   $Q^*$ using  a  neural network $Q(s,a; w) \approx Q(s,a)$, where $w $ is the network weights. In the $t$-th iteration, the DQN updates the weights using the gradient of the least squares loss function
\#\label{eq:dqn_loss}
L_t(w) = \bigl\{Q(s_t,a_t; w) -  \bigl  [ r_t + \gamma \max_{a' \in \cA} Q(s_{t+1},a' ; w_t)\bigr ] \bigr \} ^2.
\#
A variety of extensions are proposed to improve over DQN, including Double DQN \citep{hasselt2016deep}, dueling DQN \citep{wang2016dueling}, bootstrap DQN \citep{osband2016deep}, asynchronous DQN \citep{mnih2016asynchronous}, averaged-DQN \cite{anschel2017averaged} and prioritized experience replay \citep{schaul2016prioritized}.

In addition to the value-based methods, the policy-based methods directly model the optimal policy. 
The  objective of policy-based methods is to find a policy that maximizes the expected reward of a stochastic policy $\pi_{\theta}$ parametrized by  $\theta\in\Theta$.

The policy gradient methods aims at finding a weight $\theta  $ that maximizes  $  J(\pi_{\theta})$ via gradient descent. The stochastic policy gradient theorem  \citep{sutton2000policy} states that
\#\label{eq:SPGT}
\nabla _{\theta} J(\pi_{\theta} ) = \mathop{\EE}_{s,a}
\left [ \nabla _{\theta} \log \pi _{\theta} (a | s) Q^{\pi_{\theta}}(s,a) \right ].
\#
The REINFORCE algorithm \citep {williams1992simple} updates $\theta$ using $\nabla _{\theta} \log \pi_{\theta} (a_t | s_t) \cdot R_t $. Moreover, the actor-critic methods, \citep{konda2000actor} and vanilla A3C  \citep{mnih2016asynchronous} use neural network $Q(s, a; w )$ or $V(s; w )$ to estimate the value function $Q^{\pi_{\theta}}(s, a) $ associated to policy $\pi_{\theta}$. This algorithm combines the value-based and policy-based perspectives together, and  is recently  used to  achieve superhuman performance in the game of \textit{Go} \cite{silver2017mastering}.

\subsection{Reinforcement Learning Methods for Continuous Action Space}
When the action space is continuous, value-based methods will no longer be computationally tractable because of taking  maximum over the action space $\cA$ in \eqref{eq:dqn_loss}, which in general cannot   be computed efficiently. The reason is that the neural network $Q(s, a; w)$ is  nonconvex when viewed as a function of $a$;  $\max _{a \in \cA } Q(s, a; w)$ is   the global minima of a nonconvex function, which is NP-hard to obtain in the worst case.

To address this issue, the continuous Q-learning \citep{gu2016continuous} approximates action value function $Q(s,a)$ by neural networks 
\$Q(s, a; \theta^V, \theta^A ) = V(s; \theta^V) + A(s, a; \theta^A),\$
where $A(s, a; \theta^A)$ is further parameterized as a quadratic function w.r.t $a$. So the maximization over $a$ has analytic solution.

Moreover, it is also possible to adapt  policy-based methods to continuous  action spaces by considering deterministic policies $ \mu_{\theta}\colon \cS \rightarrow \cA$. Similar to \eqref{eq:SPGT},   the  deterministic policy gradient (DPG) theorem
\citep{silver2014deterministic} states that
\#\label{eq:DPGT}
\nabla _{\theta} J(\mu_{\theta} ) = \mathop{\EE}_{s \sim \rho^{\mu_{\theta}} } \left [ \nabla _{\theta} \mu _{\theta} (s) \nabla_{a} Q^{\mu_{\theta}}(s,a)|_{a=\mu_{\theta}(s)} \right ].
\#
Furthermore, this deterministic version of the policy gradient theorem can be viewed as the limit of \eqref{eq:SPGT} with the variance of $\pi_{\theta}$ going to zero. Based on \eqref{eq:DPGT}, the DPG algorithm \citep{silver2014deterministic}  and the DDPG  algorithm  \citep{lillicrap2015continuous} are proposed.
A related line of work is policy optimization methods, which improve the policy gradient method using  novel  optimization techniques. These methods include  natural  gradient descent \citep{kakade2002natural}, trust region optimization \citep{schulman2015trust}, proximal gradient descent \citep{schulman2017proximal}, mirror descent \citep{montgomery2016guided}, and entropy regularization \citep{o2016pgq}.

\subsection{Reinforcement Learning Methods for Hybrid Action Space}\label{sec-hyb-ap}
A related body of literature is the recent work on reinforcement learning with a structured action space, which  contains finite actions each parametrized by a continuous parameter.

To handle such parametrized actions, \citet{hausknecht2016deep} applies the DDPG algorithm on the relaxed action space \eqref{eq:relaxed_action_sp} directly. A More reasonable approach is to update the discrete action and continuous action separately with two different methods. 

\citet{masson2016reinforcement} propose a learning framework that alternately updates the network weights for discrete actions using Q-learning (Sarsa) and for continuous parameters using policy search (eNAC). Similarly, \cite{khamassi2017active} uses Q-learning for discrete actions and policy gradient for continuous parameters. These two methods both need to assume a distribution of continuous parameters and are both on-policy.

\section{Parametrized Deep Q-Networks (P-DQN)}

This section introduces the proposed framework to handle the application with hybrid discrete-continuous action space.
We consider an MDP with a parametrized action space  $\cA$ defined in \eqref{eq:p_action_sp}. For $a\in \cA$, we denote the action value function by $Q(s, a) = Q(s, k, x_k)$ where $s \in \cS$, $k \in [K]$, and $x_k \in \cX_k$. Let $k_t$ be the discrete action selected at time $t$ and let $x_{k_t}$ be the associated continuous parameter. Then   the Bellman equation becomes
\#\label{eq:bellman_param}
Q (s_t,k_t, x_{k_t}) = \mathop{\EE}_{r_t,s_{t+1}} \Bigl[ r_t + \gamma \max_{k \in [K] } \sup_{x_k \in \cX_k} Q (s_{t+1}, k, x_k) \Big| s_t = s, a_t = (k_t, x_{k_t})\Bigr ].
\#
    Here inside the conditional expectation   on the right-hand side of \eqref{eq:bellman_param},  we first solve $x_k^* = \argsup_{x_k \in \cX_k}  Q(s_{t+1} , k, x_k)$   for each $k \in [K]$, and then take the largest
    $Q(s_{t+1}, k, x_k^*)$. Note that taking supremum over  continuous space $\cX_k$ is computationally intractable.  However, the right-hand side of  \eqref{eq:bellman_param} can be evaluated efficiently providing $x_k^*$ is given.

    To elaborate this idea, first note that, when the function $Q$ is fixed,  for any $s \in \cS$ and $k\in [K]$, we can view $\argsup_{x_k \in \cX_k} Q(s, k, x_k)$
 as  a function    $x_k^Q\colon  \cS\rightarrow \cX_k.$  
 Then we can rewrite the Bellman equation in \eqref{eq:bellman_param} as
 \$ 
Q (s_t,k_t, x_{k_t}) = \mathop{\EE}_{r_t,s_{t+1}} \Bigl [
  r_t + \gamma \max_{k \in [K] }  Q \bigl (  s_{t+1}, k, x_k^Q(s_{t+1}) \bigr )  \Big| s_t = s \Bigr ] .
    \$
Note that this new Bellman equation resembles the classical Bellman equation in \eqref{eq:bellman} with $\cA = [K]$. Similar to the deep Q-networks, we use a deep neural network $Q(s, k, x_k; \omega)$ to approximate $Q (s , k, x_k)$, where $\omega$ denotes the network weights.  Moreover, for such a $Q(s, k, x_k; \omega)$, we approximate  $x_k^{Q}(s)$ 
with  a deterministic policy network $x_k(\cdot ; \theta ) \colon \cS \rightarrow \cX_k$, where $\theta$ denotes  the network weights of the policy network. That is, when $\omega$ is fixed, we want to find $\theta$ such that
\#\label{eq:theta_cond}
Q\bigl ( s, k, x_k(s; \theta) ; \omega \bigr  ) \approx  \sup_{x_k \in \cX_k} Q(s, k, x_k; \omega)~ ~~\text{for each}~ k \in [K].
\#

Then similar to DQN,  we could estimate $\omega$ by minimizing the mean-squared Bellman error via gradient descent. In specific, in the $t$-th step,  let $\omega_t$ and $\theta_ t$ be the weights of the value network and the deterministic policy network, respectively.
To incorporate multi-step algorithms, for a fixed  $n \geq 1$, we define  the $n$-step target  $y_t$ by
\# \label{eq:target}
y_t =  \sum_{i=0}^{n-1} \gamma^{i}  r_{t+i } + \gamma ^n    \max_{k \in [K]} Q\bigl ( s_{t+n}, k, x_k(s_{t+n}, \theta _t); \omega_t\bigr ).
\#
We use the least squares loss function for $\omega$ like DQN. Moreover,  since we aim to find $\theta  $ that maximize  $Q \bigl ( s, k, x_k(s; \theta ) ; \omega  \bigr )$ with $\omega$ fixed,  we use the loss function for $\theta  $ as following
\# \label{eq:loss_omega}
\ell_{t}^Q (\omega) =  \frac{1}{2}  \bigl [ Q \bigl ( s_t , k_t, x_{k_t}; \omega \bigr )  - y_t  \bigr ] ^2  ~~ \text{and}~~\ell_t ^\Theta (\theta  ) = - \sum_{k =1}^K Q\bigl ( s_t, k, x_k( s_t; \theta ) ; \omega_t \bigr )
\#

By \eqref{eq:loss_omega}  we   update the weights using   stochastic gradient methods. In the ideal case, we would minimize the loss function $\ell_t^{\Theta} (\theta )$ in \eqref{eq:loss_omega}  when $\omega_t$ is fixed. From the results in stochastic approximation methods \citep{kushner2006stochastic}, we could approximately achieve such a  goal in an online fashion via a two-timescale update rule \citep{borkar1997stochastic}. In specific, we update $\omega$ with a stepsize $\alpha_t$  that  is asymptotically negligible compared with the stepsize $\beta_t$ for $\theta $. In addition, for the validity of stochastic approximation, we   require $\{ \alpha_t, \beta_t \} $ to satisfy the Robbins-Monro condition \citep{robbins1951stochastic}.
We present the P-DQN algorithm with experienced replay  in Algorithm \ref{algo:main1}.

\begin{algorithm}[h!]
\caption{Parametrized Deep Q-Network (P-DQN) with Experience Replay}\label{Alg-PDQN-RB}
\begin{algorithmic}
\small
\STATE{{\bf Input:}  Stepsizes $\{\alpha_t, \beta_t\}_{t\geq 0}$
, exploration parameter $\epsilon$, minibatch size $B$,   
a  probability distribution   $\xi $.}
\STATE{Initialize network weights  $\omega_1$ and $\theta_1$.}
\FOR {$t = 1, 2, \ldots, T$}
	\STATE{Compute action parameters $x_k \leftarrow x_{k}(s_t, \theta_t) $.}
	\STATE{Select  action $a_t = (k_t, x_{k_t})$ according to the $\epsilon$-greedy policy	$$
	a_t  = \begin{cases} \text{a sample from distribution}~  \xi   & \hfill \text{with probability}  ~\epsilon, \\
	  (k_t, x_{k_t})~~\text{such that}~~k_t= \argmax _{k\in [K] } Q(s_{t}, k, x_{k}; \omega_t )     & ~~\text{with probability}   ~1- \epsilon.
	\end{cases}$$}
    \STATE{Take action $a_t$, observe reward $r_t$ and the next state $s_{t+1}$.}
	\STATE Store transition $[s_t, a_t, r_t, s_{t+1} ]$ into $\cD$.
	
	\STATE Sample $B$ transitions $\{ s_b, a_b, r_b, s_{b+1} \}_{b\in [B]}$ randomly from $\cD$.
	\STATE {Define the target $y_b $  by	$$
	y_b = \begin{cases} 
    r_b & \hfill \text{if }s_{b+1}~\text{is the terminal state,}\\
	r_b + \max_{k \in [K]}  \gamma  Q\bigl ( s_{b+1}, k , x_k(s_{b+1}, \theta_t);   \omega_t \bigr ) & ~\hfill \text{if }~\text{otherwise.}
	\end{cases}
	$$}
	\STATE{Use data $\{ y_b, s_b, a_b \}_{b \in [B]} $ to compute the stochastic gradient $\nabla_{\omega} \ell_t^Q( \omega)$ and $\nabla _{\theta} \ell_t^{\Theta} (\theta)$.}
	\STATE{Update the weights by $\omega_{t+1} \leftarrow \omega_t - \alpha_t \nabla_{\omega} \ell_t^Q( \omega_t)$ and  $\theta_{t+1}  \leftarrow \theta_{t} - \beta_t   \nabla _{\theta} \ell_t^{\Theta} (\theta_t)$. }
\ENDFOR
 \end{algorithmic}
\label{algo:main1}
\end{algorithm}

Note that  this algorithm  requires a distribution $\xi$ defined on the action space $\cA$ for exploration.
In practice, if each $\cX_k$ is a compact set in the Euclidean space (as in our case), $\xi$ could be defined as the uniform distribution over $\cA$. In addition, as in the DDPG algorithm \citep{lillicrap2015continuous}, we  can also add additive noise to the continuous part of the actions for exploration. Moreover, we use experience replay \citep{mnih2013playing} to reduce the dependencies among the samples, which can be replaced by more sample-efficient methods such as prioritized replay \citep{schaul2016prioritized}.

Moreover, we note that our P-DQN algorithm can   easily incorporate  asynchronous gradient descent to speed up the training process.  Similar to the asynchronous $n$-step DQN \cite{mnih2016asynchronous}, we consider a centralized distributed training framework where each process
computes its local gradient and communicates with a global ``parameter server''. In specific, each local process runs an independent game environment to  generate transition  trajectories and uses its own transitions to compute  gradients with respect to
$\omega$ and $\theta$.   These local gradients are   then aggregated across   multiple processes  to update the
global parameters. 
Aggregating independent stochastic gradient  decreases the variance of gradient estimation, which yields better algorithmic stability. We present the  asynchronous $n$-step P-DQN algorithm in Algorithm \ref{alg:dpqn} in Appendix. For simplicity, here we only describe the algorithm for each local process, which fetches $\omega$ and  $\theta$ from the parameter server and computes the gradient. The parameter server stores the global parameters $\omega$, $\theta$, and update them using the gradients sent from the local processes.

\begin{remark}
The key differences between the methods in \ref{sec-hyb-ap} and P-DQN are as follows.

\begin{itemize}
\item In \citet{hausknecht2016deep}, the discrete action types are parametrized as some continuous values, say $f$. And the discrete action that is actually executed is chosen via $k = \arg\max_i f(i)$ or randomly with probability $\text{softmax}(f)$. Such a trick actually relaxes the hybrid action space into a continuous action space, upon which the classical DDPG algorithm can be applied. However, in our framework, the discrete action type is chosen directly by maximizing the action's $Q$ value explicitly as illustrated in Figure \ref{fig:network}.
\item \citet{masson2016reinforcement} and \cite{khamassi2017active} use on-policy update algorithm for continuous parameters. The $Q$ network in \citet{hausknecht2016deep} is also an on-policy action-value function estimator of current policy ($Q^\pi$) if the discrete action is chosen via $\text{softmax}(f)$. While P-DQN is an off-policy algorithm.
\item Note that P-DQN can use human players' data, while it is hard to use human players' data in \citet{hausknecht2016deep} because there is only discrete action $k$ without parameters $f$.

\end{itemize}

\begin{figure*}
 \begin{center}
\begin{tabular}{cc}
   \includegraphics[width=.4\textwidth,angle=0, bb= 0 0 439 400]{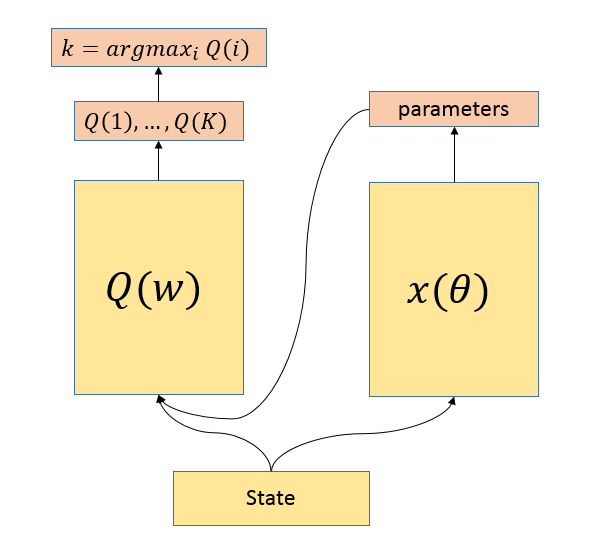}
      &\includegraphics[width=.4\textwidth,angle=0, bb= 0 0 439 400]{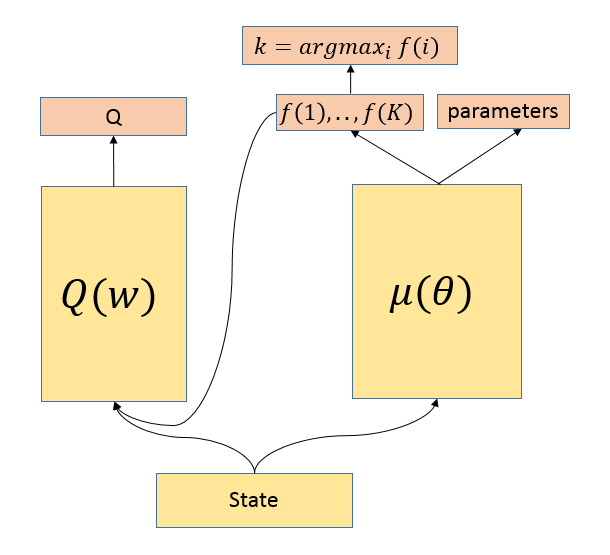}\\[0pt]
      (a) Network of P-DQN     & (b) Network of DDPG\\
\end{tabular}
\end{center}
\vskip-10pt
\caption{Illustration of the networks of P-DQN and DDPG \citep{hausknecht2016deep}. P-DQN selects the discrete action type by maximizing $Q$ values explicitly; while in DDPG \citep{hausknecht2016deep},
the discrete action is chosen via $\arg \max_i f_i$ or randomly with probability $\text{softmax}(f)$, where $f$ can be seen as a continuous parameterization of $K$ discrete action types. 
Note, more complexed structure can be designed in the Q-network of P-DQN in order to structure the parameterized relation between $Q(k)$ and $x(k)$. In the following experiments, we just input all the parameters into every $Q(k)$, which actually means all the discrete actions share the whole continues parameters.}
\label{fig:network}
\end{figure*}
\end{remark}

\subsection{The Asynchronous $n$-step P-DQN Algorithm}

Similar to the asynchronous $n$-step DQN in \cite{mnih2016asynchronous}, we can use asynchronous $n$-step P-DQN algorithm to speed up the training process. We present the  asynchronous $n$-step P-DQN algorithm in Algorithm \ref{alg:dpqn}. 
Notice when $n>1$, $n$-step DQN or $n$-step P-DQN is no longer an off-policy algorithm. However $n$-step bootstrap tactic can improve the convergence speed for delayed-reward or long-episode reinforcement learning problem.

\begin{algorithm}
\caption{The Asynchronous P-DQN Algorithm }\label{Alg-Async}
\begin{algorithmic}
\small
\STATE{{\bf Input:}  exploration parameter $\epsilon$,     a   probability distribution   $  \xi $ over the action space $\cA$    for exploration, the max length of multi step return $t_{\max}$, and maximum number of iterations $N_{\text{step}}$.}
\STATE Initialize global shared parameter $\omega$ and $\theta$
\STATE Set global shared counter $N_\text{step}=0$
\STATE Initialize local step counter $t \gets 1$.
\REPEAT
\STATE Clear local gradients $\ud\omega \gets 0$, $\ud\theta \gets 0$.
\STATE $t_\text{start} \leftarrow  t$
\STATE Synchronize local parameters  $   \omega' \leftarrow \omega$ and  $  \theta' \leftarrow  \theta$ from the parameter server.
\REPEAT

\STATE{Observe state $s_t$ and let $x_k  \leftarrow x_k(s_t, \theta') $}
\STATE{Select  action $a_t = (k_t, x_{k_t})$ according to the $\epsilon$-greedy policy	$$
	a_t = \begin{cases} \text{a sample from distribution}~  \xi   &\text{with probability}  ~\epsilon, \\
	  (k_t, x_{k_t})~~\text{s.t.}~~k_t= \argmax _{k\in [K] } Q(s_t, k, x_{k}; \omega' )     & \text{with probability}   ~1- \epsilon.
	\end{cases}$$}
\STATE{Take action $a_t $, observe reward $r_{t}$ and the next state $s_{t+1}$.}
\STATE $t \gets t + 1$
\STATE $N_\text{step} \gets N_\text{step} + 1$
\UNTIL   $s_t$ is the terminal state \textbf{or} $t - t_{\text{start}} =  t_{\max}$
\STATE Define the target \\
$y =    \left\{
    \begin{array}{l l}
      0  \quad & \text{for terminal } s_t\\
      \max_{k\in [K] }Q[s_t,k, x_{k}(s_t, \theta') ;\omega'] \quad & \text{for non-terminal } s_t
    \end{array} \right.$
\FOR {$i = t-1,\ldots,t_\text{start} $}
\STATE $y \gets r_i + \gamma\cdot y$
\STATE Accumulate gradients: $\ud\theta \gets \ud\theta + \nabla_\theta \ell^{\Theta}_t(\theta')$, $\ud\omega \gets \ud\omega + \nabla_\omega \ell_{t}^Q(\omega')$
\ENDFOR
\STATE Update global $\theta$ and $\omega$ using $\ud\theta$ and $\ud\omega$ with RMSProp (\cite{hinton2012neural}).
\UNTIL $N_\text{step} > N_\text{max}$
\end{algorithmic}
\label{alg:dpqn}
\end{algorithm}

\section{Experiments} 

We validate the proposed P-DQN algorithm in 1) a simulation example, 2) scoring a goal in simulated RoboCup soccer and 3) the solo mode in game KOG.

To evaluate the performance, we compared our algorithm with \citet{hausknecht2016deep} and DQN under fair condition for all three scenarios. \citet{hausknecht2016deep} seems the only off-policy method we are aware that solves the hybrid action space problem with deterministic policy, which can be estimated more efficiently compared with stochastic policies. DQN with discrete action approximation is also compared in the simulation example. In DQN and P-DQN, we use a dueling layer to replace the last fully-connected layer to accelerate training.

\subsection{A Simulation Example}

Suppose there is a squared plate in the size $2\times2$. The goal is to ``pull'' a unit point mass into a small target circle with radius $r=0.1$. In each unit time $\Delta t = 0.1$, a unit force $F$, i.e. $|F|=1$, with constant direction can be applied to the point mass or a soft ``brake'' can be used to reduce the velocity of point mass by $0.1$ immediately. The effect of force $F$ follows the Newton mechanics and the plate is frictionless.

Let the coordinate of point mass and the target circle center be $x \in \RR^2, y\in \RR^2$, respectively. The state is represented as an 8-dim vector $s = (x, \dot{x}, y, d(x,y), 1_{d(x,y)<r})$.
The action space is $A = \{(brake),~ (pull,\theta)\}$ and the reward is given by $r_t = d(x_t,y) - d(x_{t+1},y) + 1_{goal}$. The episode begins with random $x$, $y$ and terminates if the point mass stops in the circle or runs out the square plate or the episode length exceeds 200.

To deal with the periodic problem of the  direction of movement, we use $(\cos(\alpha), \sin(\alpha))$ to represent all the direction and learn a normalized two-dimensional vector instead of a degree (in practice, we add a normalize layer at the end to ensure this). The following two experiment also use this transformation.

\begin{figure*}
 \begin{center}
\begin{tabular}{cc}
\includegraphics[width=.45\textwidth,angle=0]{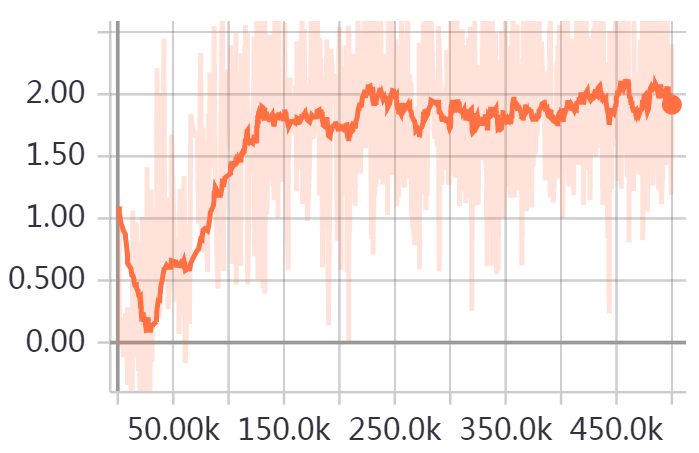}
      &\includegraphics[width=.45\textwidth,angle=0]{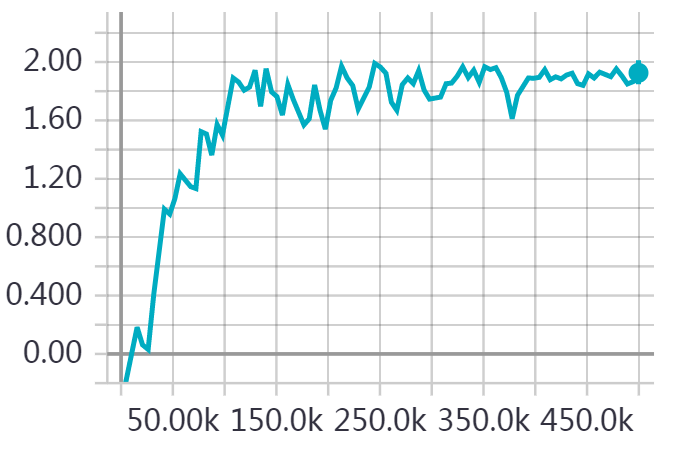}\\[0pt]
      (a) Episode reward vs. Iteration in training     & (b) Mean episode reward vs. Iteration in test\\
\end{tabular}
\end{center}
\vskip-10pt
\caption{Performance of P-DQN in Simulation Example. (a) The learning curves for P-DQN in training. We smooth the original noisy curves (plotted in light colors) to their running average (plotted in dark colors). The proposed algorithm P-DQN converged in less than $150k$ Iterations. (b) Mean episode reward of 100 trials in test. $\epsilon$-greedy exploration is removed in test.}
\label{fig:learning-curve-sim}
\end{figure*}

\begin{figure*}
	\centering
	\begin{tabular}{ccc}	\hskip-10pt\includegraphics[width=0.32\textwidth]{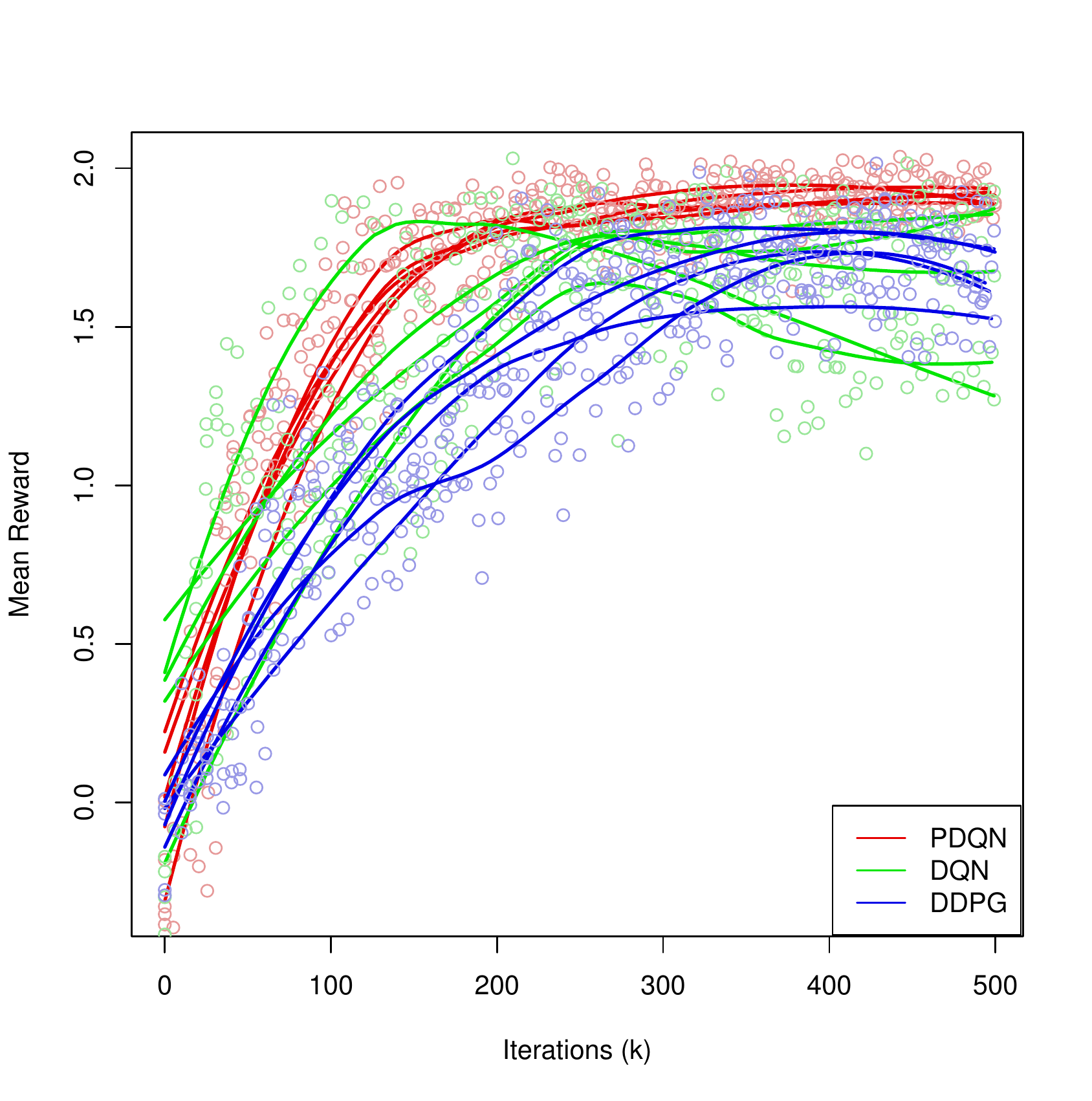}&\hskip-5pt\includegraphics[width=0.32\textwidth]{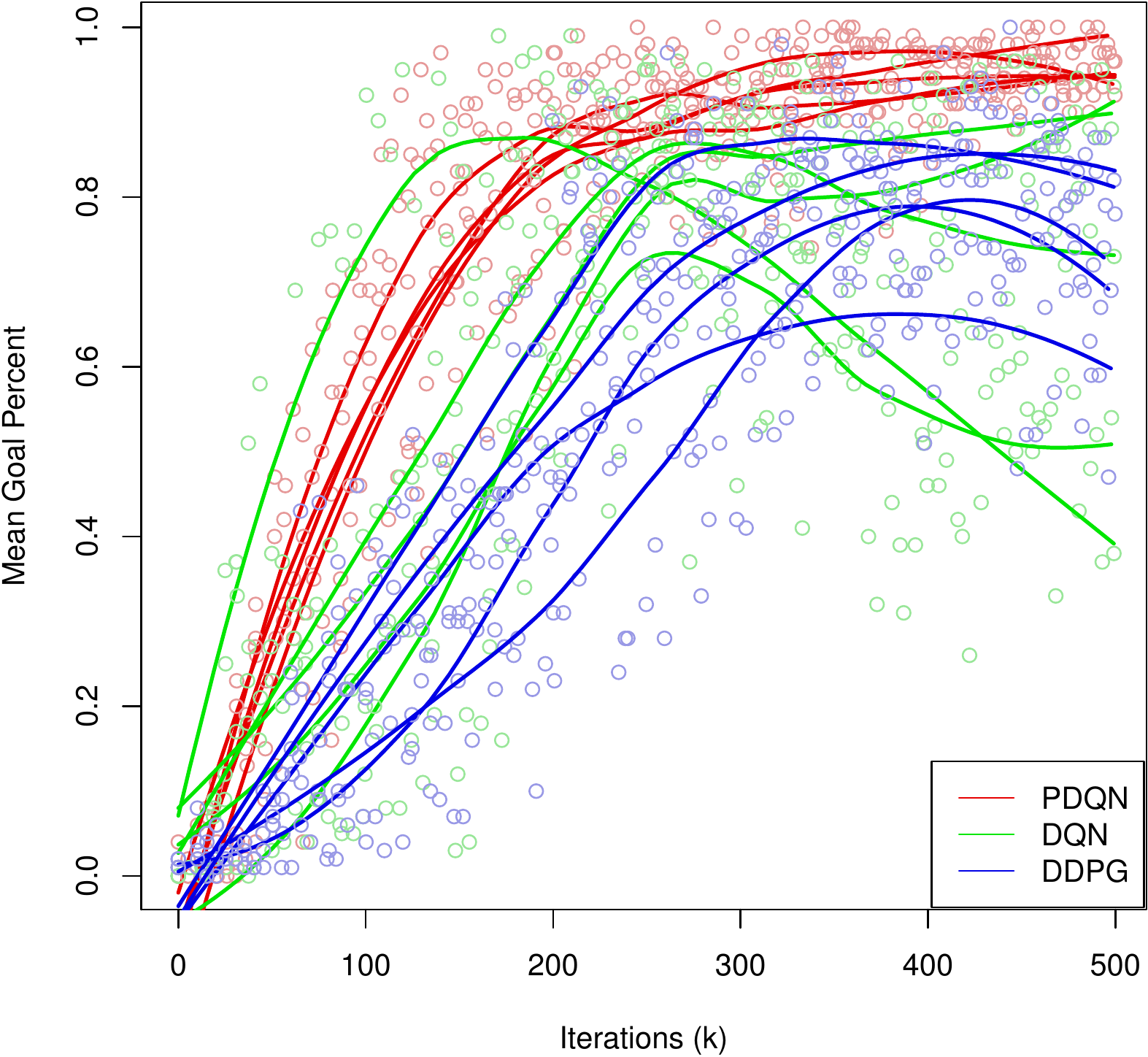}&
	\hskip-5pt\includegraphics[width=0.32\textwidth]{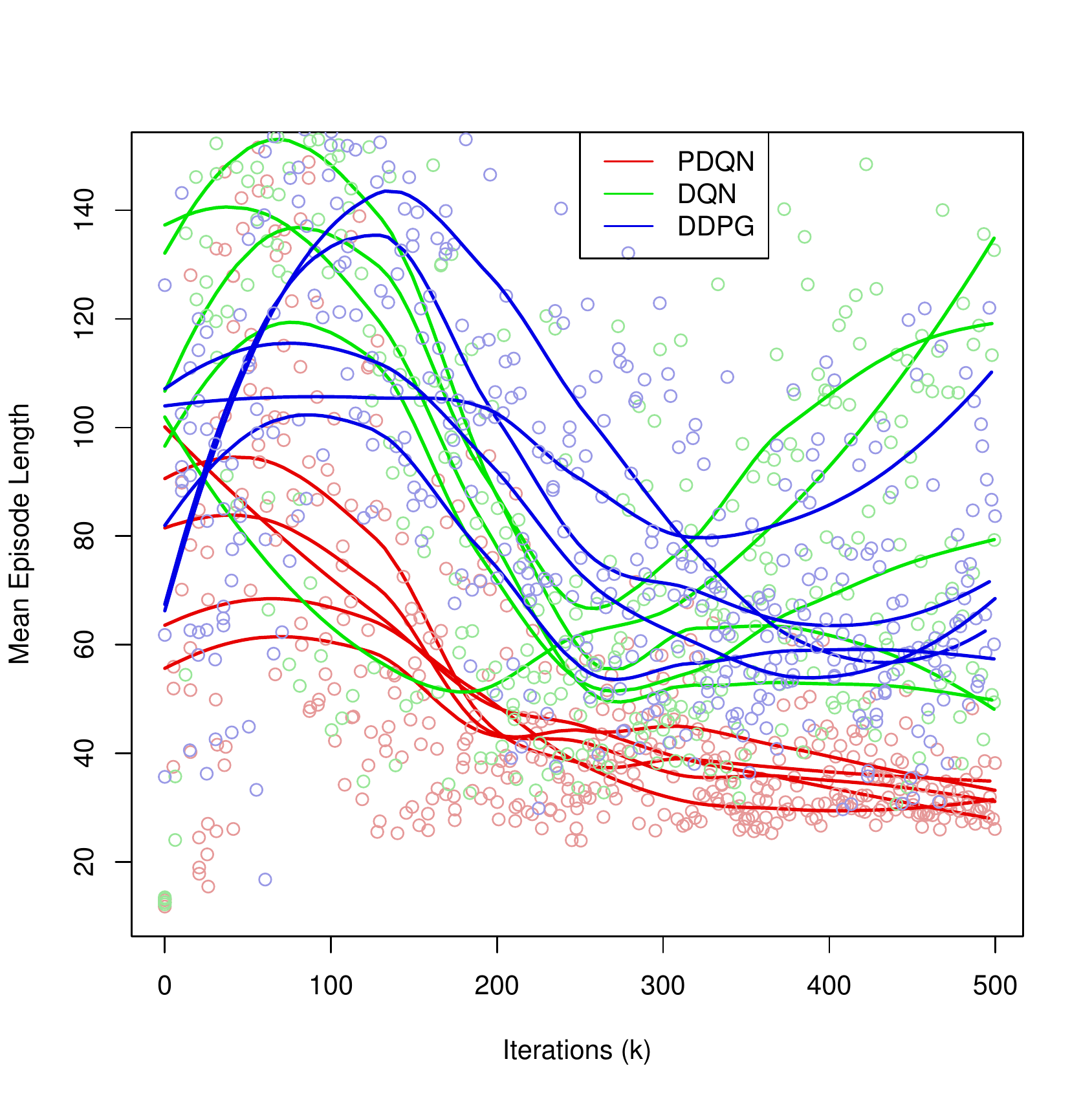} 
	\end{tabular}
	\vskip-8pt
	\caption{Comparison of three algorithms in Simulation Example. Each algorithm independently trains 5 agents and each dot (plotted in light colors) means an evaluation with average of 100 trials.  Then a smoothed curve (plotted in dark colors) is fitted to the points of each agent. The proposed algorithm P-DQN converges fast and stably to a better policy.}
	\label{fig:compare-sim}
\end{figure*}

We compare the proposed P-DQN with DDPG architecture using the same network hidden layer size. We also compared with DQN in discrete action space with $8$-direction ``pull'' and ``brake'' for completeness. We independently train 5 models for each method, and evaluate the performance during the training process with 100 trials' average. Figure \ref{fig:learning-curve-sim}, Figure \ref{fig:learning-curve-hfo}
is the learning curves for P-DQN. Figure \ref{fig:compare-sim} shows the evaluated performance in respect of mean reward, mean goal percent and mean episode length. P-DQN obviously converges much faster and more stable than its precedent work in our setting. DQN converges quickly but to a sub-optimal solution and suffers from high variance because of discretization of ``pull" direction. A demonstration of learned policy of P-DQN can be found at \url{goo.gl/XbdqHV}.

\subsection{HFO}
The Half Field Offense domain is an abstraction of full RoboCup 2D game. We use the same experiment settings with \citet{hausknecht2016deep}, scoring goals without goalie. So we just  simply summary the settings here and refer the reader to \citet{hausknecht2016deep} for details.

The state of HFO example is a 58 continuously-valued features derived through Helios-Agent2D’s \citep{Akiyama2010Helio} world model. It provides the relative position of several important objects such as the ball, the goal and other landmarks.
A full list of state features may be found at \url{https://github.com/mhauskn/
HFO/blob/master/doc/manual.pdf}.

The full action space for HFO is: \{Dash(power, direction), Turn(direction), Kick(power, direction)\}, where all the directions are parameterized in the range of $[-180, 180]$ degree and power in $[0, 100]$. Note moving forward is faster than sideways or backwards so turn the direction before moving is crucial for fast goal.

We also use the same hand-crafted intensive reward:
\$
r_t = d_{t}(a,b) - d_{t+1}(a,b) + \mathbb{I}_{t+1}^{kick} + 3(d_{t}(b,g) - d_{t+1}(b,g)) + 5\mathbb{I}_{t+1}^{goal}.
\$
where $d_{t}(a,b)$ and $d_{t}(b,g)$ are the distance between the ball and the agent or the center of goal respectively. $\mathbb{I}_t^{kick}$ is an additional reward for the agent at the first time it is close enough to kick the ball. $\mathbb{I}_t^{goal}$ is the final reward for a success goal.

\begin{figure*}
 \begin{center}
\begin{tabular}{cc}
\includegraphics[width=.45\textwidth,angle=0]{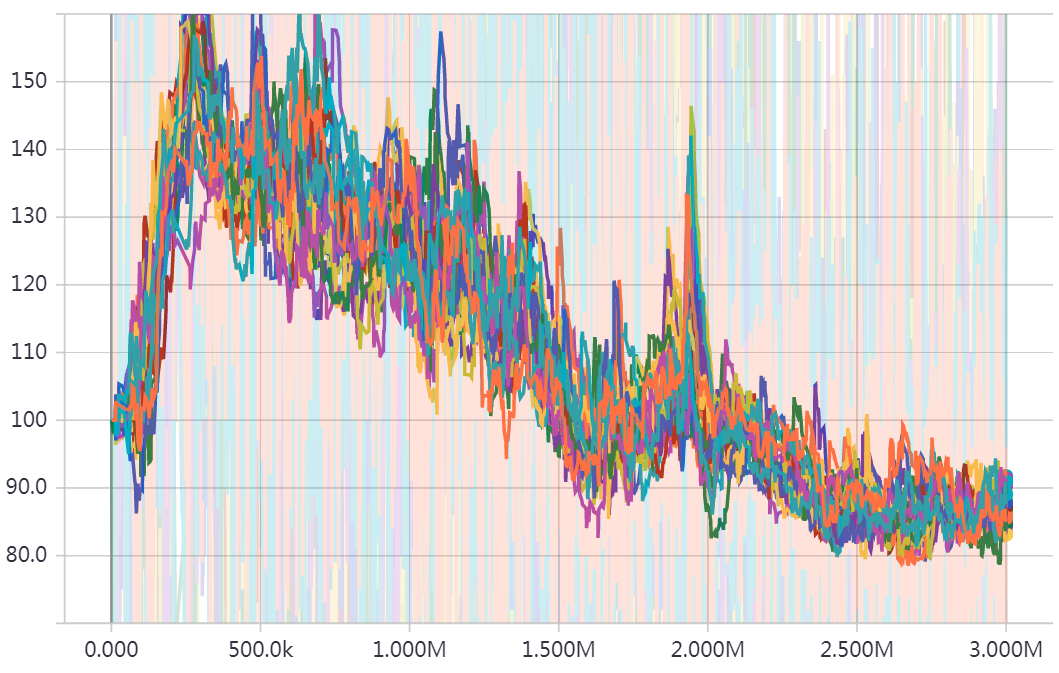}&
\includegraphics[width=.45\textwidth,angle=0]{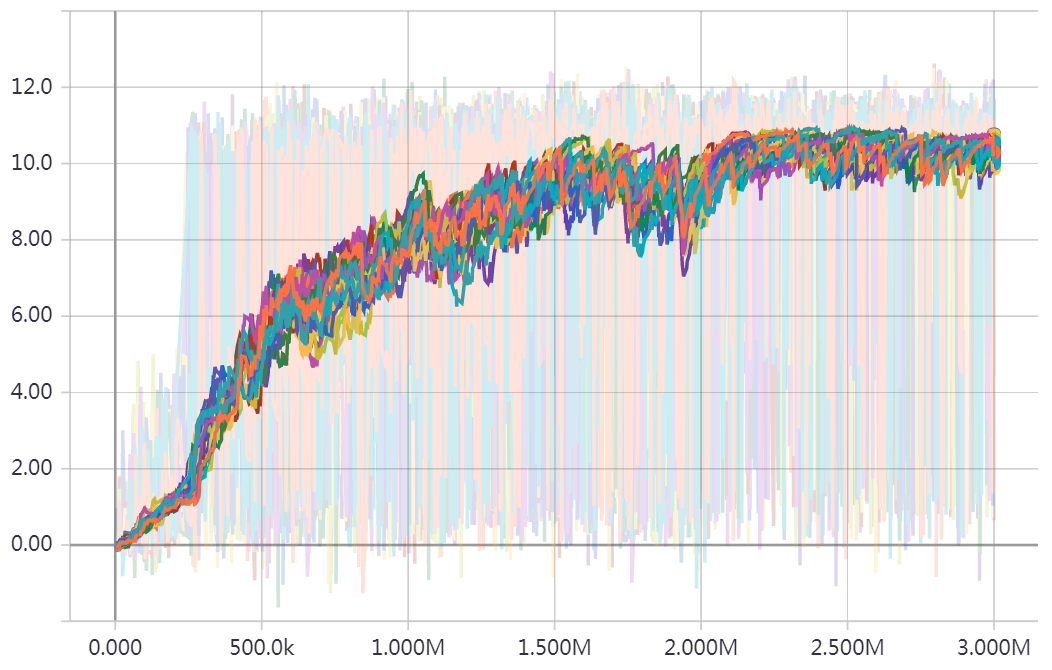}\\
(a) Episode length in training&
   (b) Episode reward sum in training\\
\end{tabular}
\end{center}
\vskip-10pt
\caption{The learning curves for P-DQN in HFO example. Different training workers are plotted
in different colors. We further smooth the original noisy curves (plotted in light colors) to their running average (plotted in dark colors). In the first 250k iterations, it learns to approach and kick the ball. The mean episode length increases because the episode is set to end if the ball is not kicked in 100 frames. After 250k iterations, it learns to goal as quick as possible as the discount factor $\gamma$ exists.}
\label{fig:learning-curve-hfo}
\end{figure*}
\begin{table}
\centering
\scriptsize
\vskip-14pt
\caption{\small Evaluation performance Comparison with different methods. The results in left column are borrowed from \citet{hausknecht2016deep}. And the performance of P-DQN is evaluated with 1000 trials. }\label{tab:eval_hfo}
{\small
\begin{tabular}{ccc|ccc}
 \toprule
  & Scoring Percent& Avg. Step to Goal& & Scoring Percent& Avg. Step to Goal\\
  \hline
Helios’& \multirow{2}{*}{.962} & \multirow{2}{*}{72.0} &\multirow{2}{*}{P-DQN$_1$}& \multirow{2}{*}{.997} & \multirow{2}{*}{78.1}  \\
 Champion &  & & & &  \\
SARSA & .81 & 70.7 &P-DQN$_2$& .997 & 78.1\\
DDPG$_1$ & 1 & 108.0 &P-DQN$_3$& .996 & 78.1 \\
DDPG$_2$ & .99 & 107.1 &P-DQN$_4$& .994 & 81.5 \\
DDPG$_3$ & .98 & 104.8 &P-DQN$_5$& .992 & 78.7 \\
DDPG$_4$ & .96 & 112.3 &P-DQN$_6$& .991 & 79.9 \\
DDPG$_5$ & .94 & 119.1 &P-DQN$_7$& .985 & 82.2 \\
DDPG$_6$ & .84 & 113.2 &P-DQN$_8$& .984 & 87.9 \\
DDPG$_7$ & .80 & 118.2 &P-DQN$_9$& .979 & 78.5 \\
\bottomrule
\end{tabular}
}
\label{tb:toy}
\end{table}

To accelerate training, we use the asynchronous version of Algorithm \ref{Alg-PDQN-RB} with 24 workers. 
Figure \ref{fig:learning-curve-hfo}
shows the learning curve of P-DQN for HFO scenario. 

Additionally, we independently trained another 8 P-DQN agents and compared the performance with the baseline results in \citet{hausknecht2016deep}. 
The result is shown in Table \ref{tab:eval_hfo}. We can see P-DQN can score more accurate and faster than DDPG with more stable performance. The training of P-DQN agent costs about 1 hour on 2 Intel Xeon CPU E5-2670 v3. In comparison, it takes three days on a NVidia Titan-X GPU to train a DDPG agent in \citet{hausknecht2016deep}. The performance video for P-DQN agent can be found at \url{https://youtu.be/fwJGR-QJ9TE}.

\subsection{\textit{Solo} mode of \textit{King of Glory}}
The game \textit{King of Glory} is the most popular mobile MOBA game in China with more than 80 million daily active players and 200 million monthly active players, as reported in July 2017. Each player controls one \textit{hero}, and the goal is to destroy the \textit{base} of the opposing team. 
In our experiments, we focus on the one-versus-one mode, which is called \textit{solo}, with both sides being the hero \textit{Lu Ban}, a \textit{hunter} type hero with a large attack range. We play against the internal AI shipped with the game.

In our experiment, the state of the game  is represented by a 179-dimensional  feature vector which is manually constructed using the output from the game engine. These features consist of two parts. The first part is  the basic attributes of the units and the second component of the features is the relative positions of other units with respect to the \textit{hero} controlled by the player as well as the attacking relations between units. We note that these features are directly extracted from the game engine without sophisticated feature engineering. We conjecture that the overall performances could be improved with a more careful  engineered set of features.

We simplify the actions of a hero into $K = 6$ discrete action types: \textit{Move}, \textit{Attack}, \textit{UseSkill1}, \textit{UseSkill2}, \textit{UseSkill3}, and \textit{Retreat}.
Some of the actions may have additional continuous parameters  to specify the precise behavior.
For example, when the  action type is  $k = \textit{Move}$, the direction of movement is given by the parameter
$x_k = \alpha$, where $\alpha \in [0, 2\pi]$. Recall that each \textit{hero}'s skills are unique.
For  \textit{Lu Ban}, the first skill is to throw a grenade at some specified location,
the second skill is to launch a missile in a particular direction,  and the last skill is to call an airship to fly in a specified direction.
A complete list of actions as well as the associated  parameters are given in Table \ref{tab:action_param}.

In KOG, the 6 discrete actions are  not always usable,
due to skills level up, lack of \textit{Magic Point (MP)}, or skills \textit{Cool Down(CD)}. In order to deal with this problem, we replace the $\max_{k \in [K]}$ with $\max_{k \in [K] \text{ and } k \text{ is usable}}$ when selecting the action to perform, and calculating multi-step target as in Equation \ref{eq:target}. 

\begin{table}
\centering
\caption{Action Parameters in KoG}\label{tab:action_param}
\vspace{5pt}
{\small
\begin{tabular}{ccc}
 \toprule
ActionType & Parameter & Description \\
  \hline
Move & $\alpha$ & Move in the direction $\alpha$ \\
Attack & - & Attack default target \\
UseSkill1 & $(x, y)$ & Skill 1 at the position $(x,y)$ \\
UseSkill2 & $\alpha$ & Skill 2 in the direction $\alpha$ \\
UseSkill3 & $\alpha$ & Skill 3 in the direction $\alpha$ \\
Retreat & - & Retreat back to our base \\
\bottomrule
\end{tabular}
}
\end{table}

\subsection{Reward for KoG}
To encourage winning the game, we adopt reward shaping, where the immediate reward takes into account Gold earned, Hero HP, kill/death, etc. Specifically, we define a variety of  statistics as follows. (In the sequel, we use subscript $0$ to represent the attributes of our side and $ 1$  to represent those of the opponent.)
\begin{itemize}
  \item Gold difference $\textit{GD} = \textit{Gold}_0 - \textit{Gold}_1$.
  This statistic measures the difference of gold gained from killing  hero, soldiers and destroying towers of the opposing team. The gold can be used to buy weapons and armors, which  enhance the
  offending and defending attributes of the hero. Using this value as the reward  encourages the hero to gain more gold.
  \item \textit{Health Point} difference ($\textit{HPD} = \textit{HeroRelativeHP}_0 - \textit{HeroRelativeHP}_1$): This statistic measures the difference of  \textit{Health Point} of the two competing heroes.
  A hero with higher \textit{Health Point} can bear more severe damages   while hero with lower \textit{Health Point} is more likely to be killed. Using this value as the reward encourages the hero to  avoid attacks and last longer before being killed by the enemy.
  \item Kill/Death $\textit{KD} = \textit{Kills}_0 - \text{Kills}_1$.  This statistic measures the historical performance of the two heroes. If a hero is killed multiple times, it  is
  usually considered  more likely to lose the game. Using this value as the reward can encourage the hero to kill the opponent and avoid  death.
  \item Tower/Base HP difference\\ $\textit{THP} = \textit{TowerRelativeHP}_0 - \textit{TowerRelativeHP}_1$,
  $\textit{BHP} = \textit{BaseRelativeHP}_0 - \textit{BaseRelativeHP}_1$. These two statistics measures the health difference  of the towers and bases of the two teams. Incorporating these two statistic in the reward encourages our hero to   attack  towers of the opposing team and  defend its own towers.
  \item Tower Destroyed $\textit{TD} = \textit{AliveTower}_0 - \textit{AliveTower}_1$. This counts  the number of destroyed towers, which rewards the  hero when it successfully destroy
  the opponent's towers.
  \item Winning Game $\textit{W} = \textit{AliveBase}_0 - \textit{AliveBase}_1$. This value indicates the winning or losing of the game.
  \item Moving forward reward: $\textit{MF} = x + y$, where $(x,y)$ is the coordinate of $\textit{Hero}_0$: This value is used as part of the reward to guide our hero to move forward
  and compete actively in the battle field.
\end{itemize}
The overall reward is calculated as a weighted sum of the  time differentiated statistics defined above. In specific, the  exact formula is
\begin{align*}
r_t=&0.5\times 10^{-5}(\textit{MF}_t-\textit{MF}_{t-1}) + 0.5(\textit{HPD}_t - \textit{HPD}_{t-1}+ \textit{KD}_t - \textit{KD}_{t-1} + \textit{TD}_t - \textit{TD}_{t-1})\\
 +& 0.001(\textit{GD}_t - \textit{GD}_{t-1}) + (\textit{THP}_t - \textit{THP}_{t-1} + \textit{BHP}_t - \textit{BHP}_{t-1}) + 2\textit{W}.
\end{align*}
The coefficients are set roughly inversely proportional to the scale of each statistic. We note that our algorithm is not very sensitive to the change of these coefficients in a reasonable range.

\begin{figure*}
 \begin{center}
\begin{tabular}{cc}
   \includegraphics[width=.45\textwidth,angle=0]{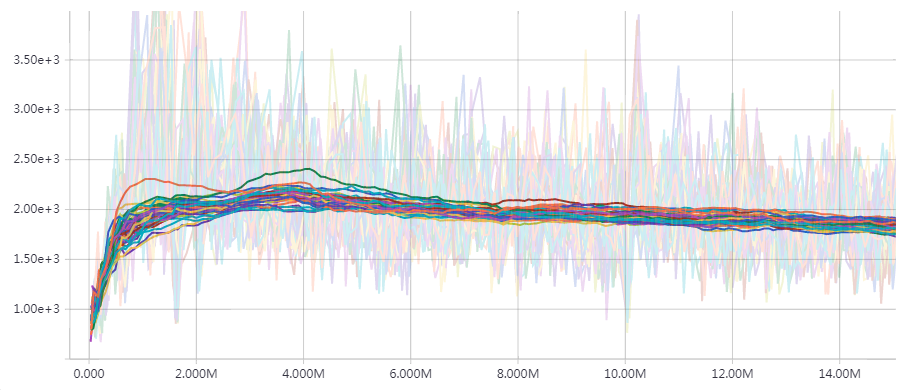}
      &\includegraphics[width=.45\textwidth,angle=0]{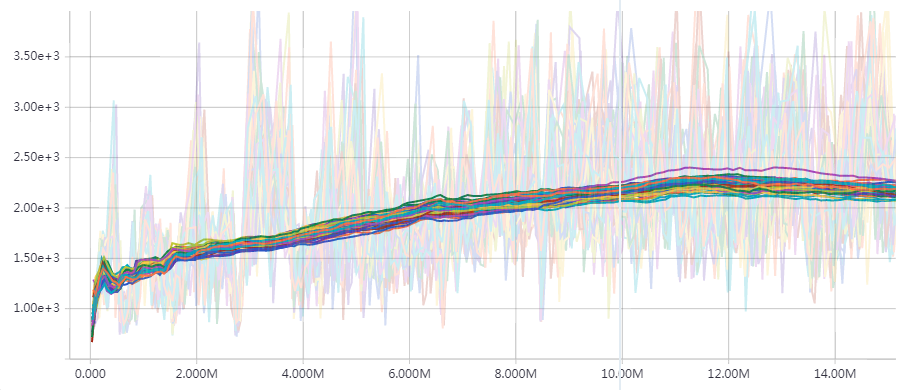}\\[0pt]
      (a1) Episode length in training     & (b1) Episode length in training\\
   \includegraphics[width=.45\textwidth,angle=0]{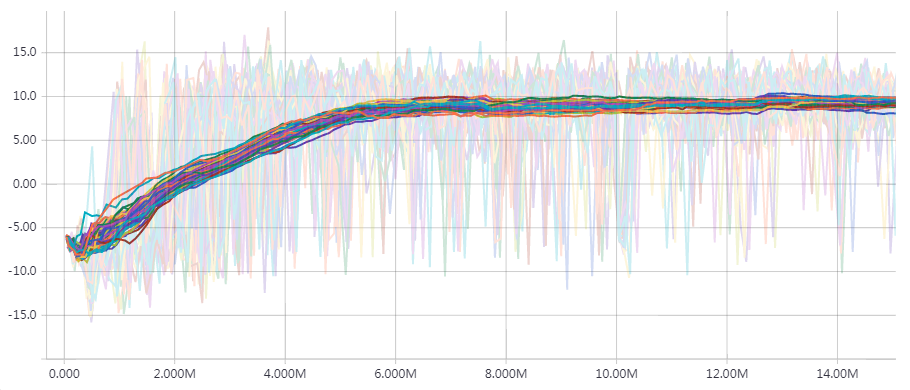}
      & \includegraphics[width=.45\textwidth,angle=0]{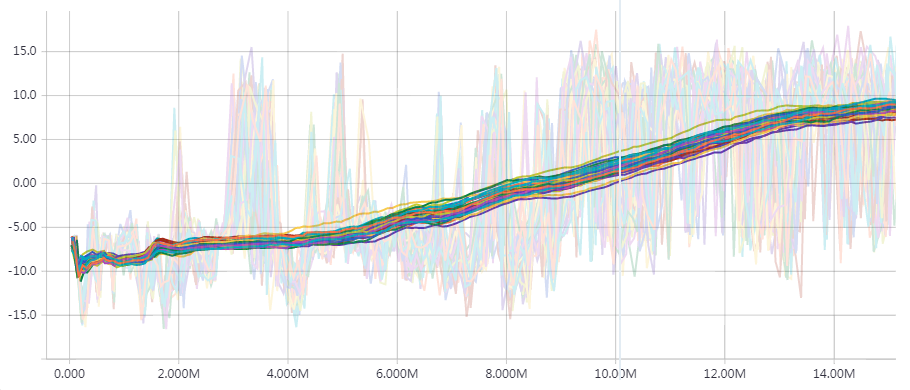}\\[0pt]
      (a2) Episode reward sum in training & (b2) Episode reward sum in training\\
   \includegraphics[width=.45\textwidth,angle=0]{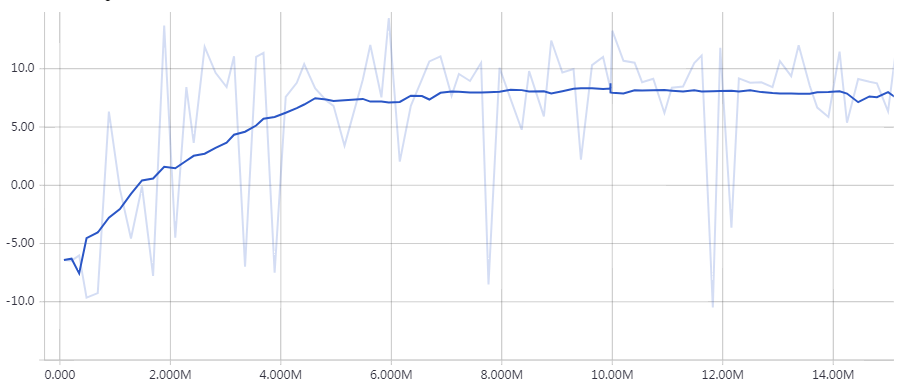}
      & \includegraphics[width=.45\textwidth,angle=0]{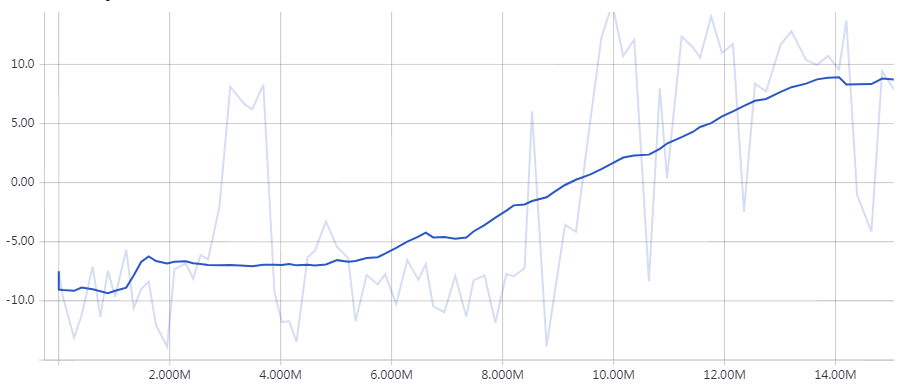}\\[0pt]
      (a3) Episode reward sum in validation  & (b3) Episode reward sum in validation\\
\end{tabular}
\end{center}
\caption{Comparison of P-DQN and DDPG for \textit{solo} games with the same hero \textit{Lu Ban}. The learning curves for different training workers are plotted
in different colors. We further smooth the original noisy curves (plotted in light colors) to their running average (plotted in dark colors). 
Usually a positive reward sum indicates a winning game, and vice versa.
(a) Performance of P-DQN. (b) Performance of DDPG}
\label{fig:learning-curve}
\vskip-5pt
\end{figure*}

We use Algorithm \ref{Alg-Async} with 48 parallel workers and frame skipping. The training and validating performances are plotted in Figure \ref{fig:learning-curve}.

From the experimental results in Figure \ref{fig:learning-curve}, we can see that our algorithm P-DQN can learn the value network and the policy network much faster comparing to \citet{hausknecht2016deep}.
In (a1), we see that the average length of games increases at first, reaches its peak when the two players' strength are close, and decreases when our player can
easily defeat the opponent. In addition,  in (a2) and (a3), we see that the total rewards in an episode increase consistently in training as well as in test settings.

\section{Conclusion}

Previous deep reinforcement learning algorithms mostly can work with either discrete or continuous action space. In this work, we consider the scenario with discrete-continuous hybrid action space. In contrast of existing approaches of approximating the hybrid space by a discrete set or relaxing it into a continuous set, we propose the parameterized deep Q-network (P-DQN), which extends the classical DQN with deterministic policy for the continuous part of actions. Several empirical experiments with comparison of other baselines demonstrate the efficiency and effectiveness of P-DQN.

\appendix
\newpage

\section{Appendix}

\subsection{More information on King of Glory}

\begin{figure*}[htp]
 \begin{center}
\begin{tabular}{ccc}
\includegraphics[width=.25\textwidth,angle=0]{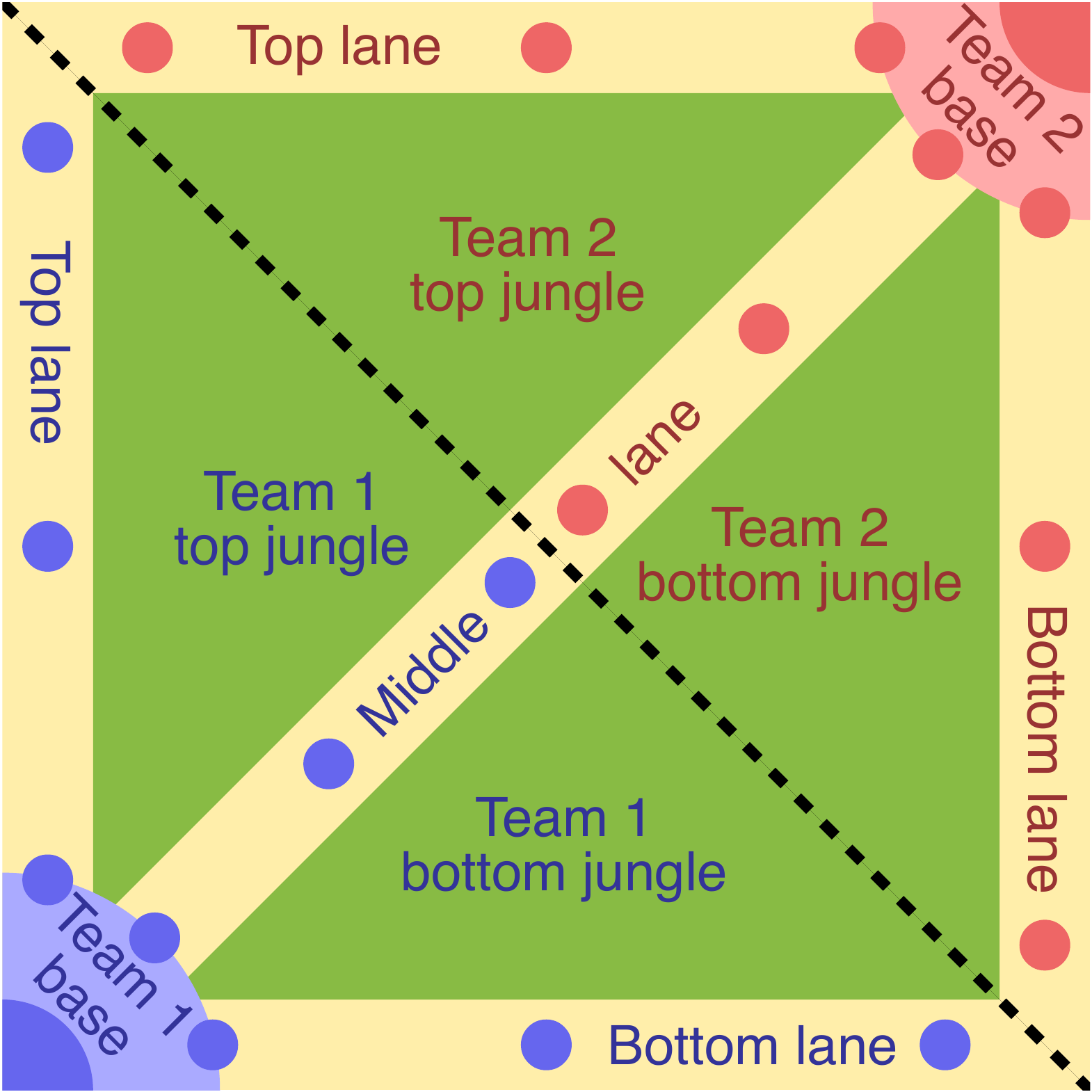}&
\includegraphics[width=.25\textwidth,angle=0]{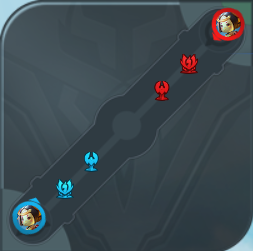}&
\includegraphics[width=.45\textwidth,angle=0]{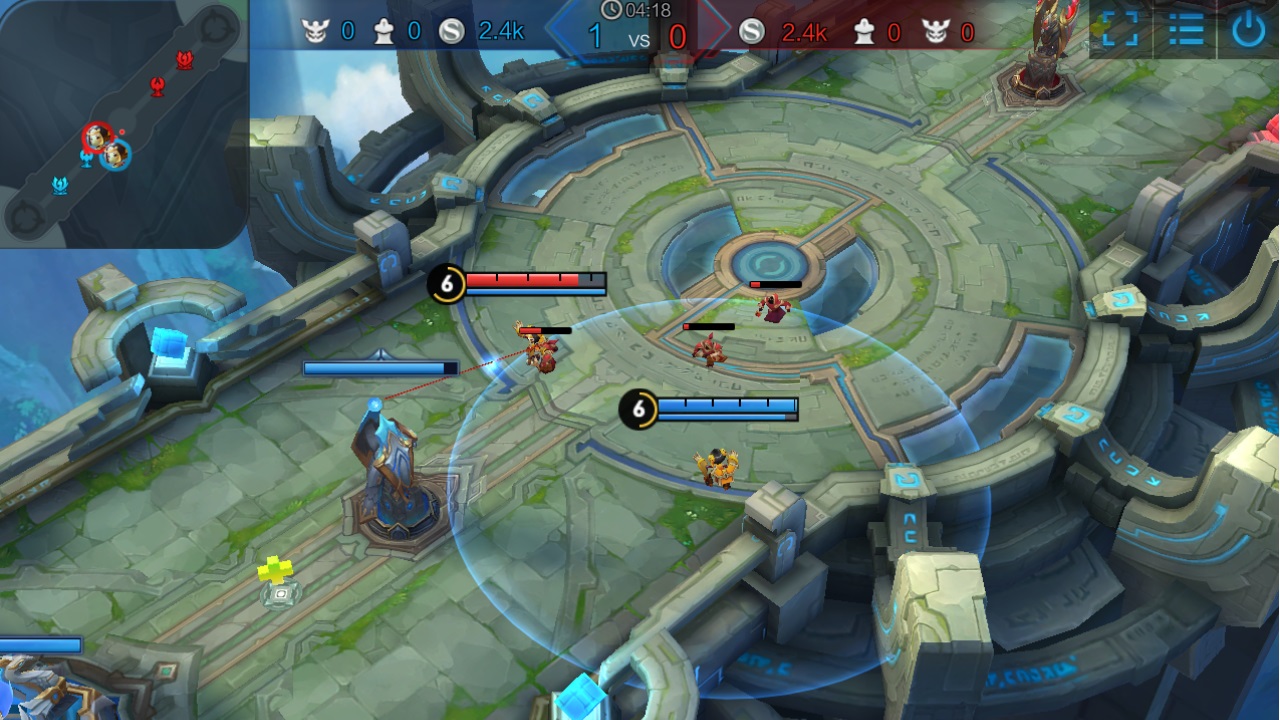}\\[0pt]
(a)   The map of a  &(b) The battle field for  
&(c) A screen shot of \\[0pt]
MOBA game & a   \textit{solo} game  & a \textit{solo} game
\end{tabular}
\end{center}

\caption{ (a) An illustration of the map of a MOBA game, where there are three lanes connecting two  \textit{bases}, with three \textit{towers} on each lane for each side.
   (b). The map of a \textit{solo} game of \textit{King of Glory}, where only the middle lane is active.    (c). A screenshot of a  \textit{solo} game of \textit{King of Glory}, where the unit under a blue bar is a \textit{hero} controlled by our algorithm and the rest of the units are the  computer-controlled units.}
\label{fig:game}
\end{figure*}

The game \textit{King of Glory} is a MOBA game,
which is  a special form of the RTS game where the players are divided into two opposing teams fighting against each other.
Each team has a team \textit{base} located in either the bottom-left or the top-right corner which are guarded by three \textit{towers} on each of the three lanes.  The \textit{towers}
can  attack the enemies when they are within its attack range.
Each player controls one \textit{hero}, which is a powerful unit that is  able to move, kill, perform skills, and purchase equipments. The goal of  the \textit{heroes} is to destroy the \textit{base} of the opposing team.
 In addition, for both teams,  there are computer-controlled units spawned periodically that march towards the opposing \textit{base} in all the three lanes. These units can  attack the enemies but cannot perform skills or purchase equipments.   An illustration of the map is in Figure \ref{fig:game}-(a), where the blue or red circles on each lane are the \textit{towers}.
 
 During game play, the  \textit{heroes} advance their levels and obtain gold by killing  units and destroying the \textit{towers}. With  gold, the \textit{heros} are able to purchase equipments such as weapons and armors to enhance their power. In addition, by upgrading to the new level, a hero is able to improve its unique skills. Whereas when a \textit{hero}  is killed by the enemy, it will 
wait for some time to reborn.

In this game, each team contains one, three, or five players.
The five-versus-five model is the most complicated mode which requires strategic collaboration among the five players.  In contrast,  the one-versus-one mode, which is called \textit{solo}, only depends on the player's control of a single \textit{hero}.  In a \textit{solo} game, only the middle lane is active; both the two players move along the middle lane to fight against each other.  The map and a screenshot of  a \textit{solo} game are given in Figure \ref{fig:game}-(b) and (c), respectively.
In our experiments, we play focus on the \textit{solo} mode. We emphasize that a typical \textit{solo} game lasts about 10 to 20 minutes where each player must make instantaneous decisions. Moreover, the players have to make different types of actions including \textit{attack}, \textit{move} and \textit{purchasing}.   Thus, as a reinforcement learning problem,  it has four main difficulties:
first,  the state space has huge capacity; second, since there are various  kinds of  actions, the action space is complicated; third,  the reward function is not well defined; and fourth, heuristic search algorithms  are not feasible since the game is in real-time. Therefore, although we consider  the simplest mode of \textit{King of Glory}, it is still a challenging game for artificial intelligence. 

\subsection{Parameter setting}

\textbf{Simulation:}
In Simulation example, we use $B = 32$ with replay memory size $10k$ in Algorithm \ref{Alg-PDQN-RB}. The network $x(\theta)$ is in size of  64-32 nodes in each hidden layer, with the Relu activation function and $Q(\omega)$ is in size of 64-32-32. A uniform sample distribution is used in $\epsilon$-greedy and $\epsilon$ is annealed from $1$ to $0.1$ over first $30k$ iterations and stay constant. The learning rate is annealed from $0.001$ to $0$.

\textbf{HFO:}
In HFO example, we use $B = 32$ and replay memory size $1k$ for each worker, and the network $x(\theta)$ is in the size of  256-128-64 nodes in each hidden layer, and $Q(\omega)$ is in size of 256-128-64-64. A uniform sample distribution is used in $\epsilon$-greedy and $\epsilon$ is annealed from $1$ to $0.1$ over first $150k$ iterations and stay constant. The learning rate is annealed from $0.001$ to $0$.Additionally, \citet{hausknecht2016deep} suggest to use Inverting Gradients to enforce the bounded continues parameters to stay in the value range. Instead of using complicated Inverting Gradients technique, we just add a square loss penalty on the out-of-range part.

\textbf{KOG:} The network $x(\theta)$ or $\mu(\theta)$ are both in the size of  256-128-64 nodes in each hidden layer, and $Q(\omega)$ is in size of 256-128-64-64. To reduce the long episode length, we set the frame skipping parameter to 2.  This  means that we  take actions every 3 frames or equivalently,  0.2 second. Furthermore, we set $t_{\max}=20$ (4 seconds) in Algorithm \ref{Alg-Async} to alleviate the delayed reward. In order to encourage exploration, we use $\epsilon$-greedy sampling in training with $\epsilon=0.255$. In specific, the first 5 type actions are sampled
with probability of $0.05$ each and the action ``Retreat'' with probability $0.005$.  Moreover, if the sampled action is infeasible, we execute the greedy policy from the feasible ones, so the effective exploration rate is less than $\epsilon$. The learning rate in training is fixed at 0.001 in training.

\newpage
\bibliography{pdqn_arxiv}
\bibliographystyle{abbrvnat}

\end{document}